\documentclass[a4paper]{styles/svproc}
\usepackage{hyperref}
\usepackage{url}
\usepackage[tracking=true]{microtype}
\usepackage{amsmath} 
\usepackage{amssymb}  
\usepackage{graphicx}
\usepackage{epsfig} 
\usepackage{mathptmx} 
\DeclareMathAlphabet{\mathcal}{OMS}{cmsy}{m}{n} 
\usepackage{algpseudocode}
\usepackage{times} 
\usepackage{amsmath} 
\usepackage{amssymb}  
\usepackage{xspace}
\usepackage{tabularx}
\usepackage{multirow}
\usepackage{adjustbox}
\usepackage[frozencache,cachedir=minted]{minted}
\setminted{fontsize=\footnotesize}
\usepackage{url}
\usepackage[dvipsnames]{xcolor}
\usepackage[font={footnotesize}]{caption}
\usepackage{subcaption}
\usepackage{booktabs} 
\usepackage{tikz}
\usetikzlibrary{positioning,shapes.geometric,arrows,fit,shapes.symbols,svg.path,tikzmark,calc}
\usepackage{booktabs}
\usepackage{textcomp}
\usepackage{listings}
\usepackage{subcaption}
\pgfsetlayers{background,main,foreground}
\usepackage{siunitx}
\usepackage[short]{optidef}
\usepackage[inline]{enumitem}

\usepackage{url}

\usepackage[
backend=biber,
style=ieee,
]{biblatex}
\bibliography{references}

\begin{document}

\mainmatter              
\title{Breathless: An 8-hour Performance \\
Contrasting Human and Robot Expressiveness }
\titlerunning{Breathless: An 8-hour Human-Robot Dance Performance}  
%
\author{Catie Cuan\inst{1}, Tianshuang Qiu\inst{2}, Shreya Ganti\inst{2}, Ken Goldberg\inst{2}}
\authorrunning{} 
%
%
\institute{Stanford\inst{1} and UC Berkeley\inst{2} \\
\email{ccuan@stanford.edu, \{ethantqiu, shreyaganti, goldberg\}@berkeley.edu}}

\maketitle              

\vspace{-0.6cm}
\begin{abstract}
This paper describes the robot technology behind an original performance that pairs a human dancer (Cuan) with an industrial robot arm for an eight-hour dance that unfolds over the timespan of an American workday. To control the robot arm, we combine a range of sinusoidal motions with varying amplitude, frequency and offset at each joint to evoke human motions common in physical labor such as stirring, digging, and stacking. More motions were developed using deep learning techniques for video-based human-pose tracking and extraction. We combine these pre-recorded motions with improvised robot motions created live by putting the robot into teach-mode and triggering force sensing from the robot joints onstage.  All motions are combined with commercial and original music using a custom suite of python software with AppleScript, Keynote, and Zoom to facilitate on-stage communication with the dancer.  The resulting performance contrasts the expressivity of the human body with the precision of robot machinery.
Video, code and data are available on the project website:  \url{https://sites.google.com/playing.studio/breathless}
\keywords{art, dance, robot kinematics, sinusoidal motions, robochoreography}
\end{abstract}

\begin{figure}[H]
\begin{center}
\vspace{-2cm}
    \includegraphics[scale=0.4]{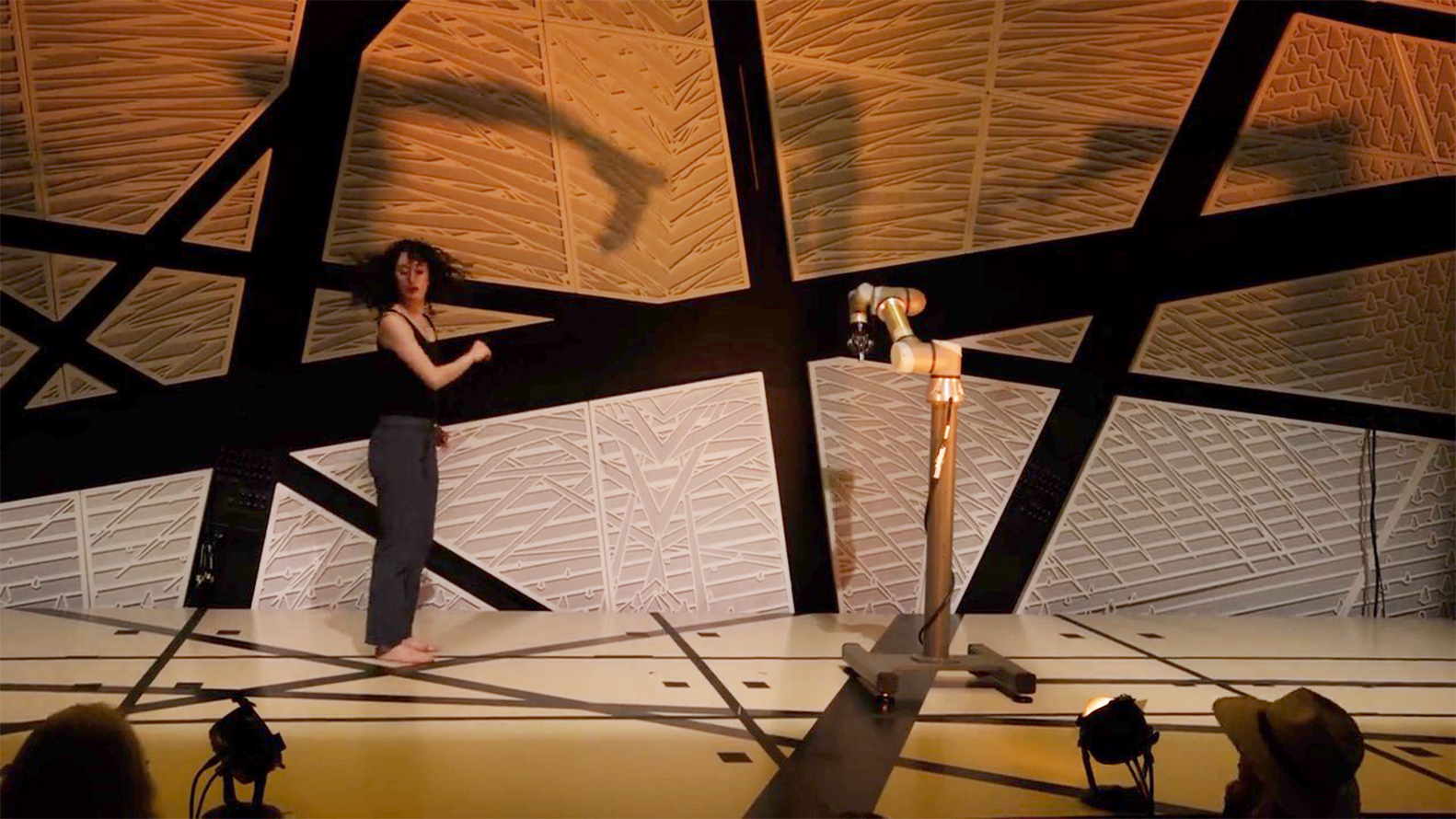}
    \caption{Catie Cuan with UR5e industrial robot arm at the National Sawdust Theater, Dec 16, 2023. }  
\end{center}
\vspace{-1cm}
\label{fig:splash}
\end{figure}

\newpage
\section{Introduction}

The combination of art and robots has a long history dating back to the origin of the word ``robot" in the theatrical play -- Rossum's Universal Robots (R.U.R.) -- written in 1920 by Czech playright Karel Capek.  In 1927, German director Fritz Lang released the silent film "Metropolis" featuring "Olympia" a female metallic robot whose name is a reference to an automaton in E.T.A. Hoffman's classic gothic novel, "Der Sandman".  American author Isaac Asimov later published "I Robot", a series of science-fiction novels that originated the Three Laws of Robotics: 
\begin{enumerate}
    \item A robot must not harm a human, or allow a human to come to harm through inaction.
    \item A robot must obey human orders, except when they conflict with the first law.
    \item A robot must protect its own existence, except when it conflicts with the first or second law.
\end{enumerate}
 The combination of artistic performance and robots also has a rich history with works by Jean Tinguely \cite{tinguelymeta}, Nam Jun Paik \cite{paik1964robot}, Survival Research Labs \cite{survivalresearchlabs1998}, Stelarc \cite{stelarc2023}, and many others.  The specific combination of dance with robots has a compelling history that includes Margo Apostolos \cite{apostolos1990robot}, William Forsythe \cite{forsythe_flags, forsythe2017}, Amy Laviers \cite{laviers2020, laviers2014style, laviers2018choreographic}, and Boston Dynamics \cite{loveme,bdmoves,uptown}. This history also includes the popular street dance style of "robot" or "popping" from the late 1960 that was popularized by Michael Jackson.   

Breathless is structured with interwoven high and low energy moments characterized by several motifs and their variations, an example being the robot and human wake-up start sequence and the faster-paced celebratory ending. To take advantage of the UR5e's full range of motion, we designed a custom minimalist mounting platform for the robot. Accessibility to all directions on stage is key to engaging audiences around the venue by storytelling with a diverse set of collaborative sequences between the UR5e and Cuan.


This project began in 2022 when we began experimenting with the OpenPose software that uses deep learning to track human body kinematics. Working with Qiu and Ganti in Goldberg's lab, we discovered that many of Cuan's natural dance motions could be represented with a series of sinusoidal motions with differing amplitude, frequency, and offset at each joint, a concept that has been used previously to determine robot joint trajectory \cite{morimoto2006, morimoto2008}.  

Sinusoidal functions of time  can be described with the equation:
\[
y(t) = A \cos(\omega t + \varphi) = A \sin(2\pi ft + \varphi)
\]

where:
\begin{itemize}
    \item $A$, amplitude, the function's largest deviation from zero.
    \item $t$, the independent variable, usually representing time in seconds.
    \item $\omega$, angular frequency, the rate of change of the function in units of radians per second.
    \item $f$, ordinary frequency, the number of oscillations (cycles) that occur per second.
    \item $\varphi$, phase in radians, the offset where the function is at $t = 0$.
\end{itemize}

In 1993, Murray and Sastry \cite{sinusoids93} showed that sinusoidal control signals at integrally related frequencies can be used to control a class of nonholonomic systems that include single-leg running robots, car-like steering robots, and especially impressively: a truck towing multiple trailers that they represent as a chained system.   

In this paper we smoothly transition between differing sinusoids at each joint to obtain smooth robot trajectories that evoke motions such as stirring, digging, stacking, and other human motions common in physical labor to evoke the history of industrialization and human work.   Additional movement motifs include: wall painting, folding laundry,  hammering,  changing a lightbulb, mopping, constructing picture frames and fulfilling orders in a warehouse. Only two props are used in the performance: a long stick that is used as a mop and paintbrush extender, and a large white sheet that evokes setting a table.

For the performance, we selected the UR5e six-axis industrial robot arm from Universal Robots, which graciously agreed to loan us two UR5e robot arms for rehearsal and production \cite{universal2008}. These robots are well-designed, precise, and robust.

This paper contributes:

1) A novel application of AI-based human pose tracking to robot motion planning using sinusoidal functions.

2) A description of custom software tools that facilitate this form of sinusoidal robot motion planning which will be made available for artists and researchers.

3) A description of the genesis of this performance, background concept, and results from the East Coast premier on 16 December 2023.

\section{Related Work}

Choreographers and roboticists have collaborated for many years. One of the earliest dances made with a robot was by Margo Apostolos \cite{apostolos1990robot} in the 1980s while a PhD student in Physical Education at Stanford University. 

Since then, several different choreographers have used robots in their work. In William Forsythe's \textit{Black Flags} \cite{forsythe_flags}, a robot recreated a dance originally choreographed for a human to perform. Gil Weinberg and collaborators' built an improvising robot musician \textit{Shimon} \cite{weinberg2009interactive} that has also performed with dancers. Daniela Rus, dance company Pilobolus, and MIT CSAIL's collaborated to make \textit{Seraph}, a dance performance with a drone. Catie Cuan and the Rad Lab used several robots in a dance performance and art installation that used both responsive and scripted elements \cite{cuan2018time}. In, Kate Ladenheim and the Rad Lab's \textit{Babyface} \cite{ladenheim2020live}, the creators built a custom wearable robot to respond to breathing. Adrienne Hart of Neon Dance worked with the University of Oxford's TORCH programme to build an interactive contemporary dance work with robots.

Robots have also appeared prominently in popular culture dances and live concert tours. Beyoncé featured robots in her recent `Renaissance' tour \cite{blistein_rollingstone}. Boston Dynamics published a blog post about choreographing their robots into dance performances and music videos \cite{danceblog}. These videos with Boston Dynamics robots were created to music by Bruno Mars \cite{uptown}, The Contours \cite{loveme}, and Katy Perry \cite{katyperry}. 

In addition to prior artistic works, human-robot interaction researchers have collaborated with choreographers, theater artists, and musicians in research settings. Guy Hoffman, et al. \cite{hoffman2014designing} noted the importance of designing expressive social robots with a strong movement foundation. Amy LaViers et al. \cite{laviers2018choreographic} described choreographic applications and techniques for robot programming and creation. A team of choreographers and roboticists from Catie Cuan and the Rad Lab designed an experimental testbed for human-robot interaction inside of dance performances and artistic installations \cite{cuan2018curtain, cuan2018time}. Human-robot co-creation continues to be explored in the dance context to study the creative and skill learning process of both agents \cite{thorn2020human, defilippo2023towards, granados2017dance}. Engineers and an artist-in-residence at Everyday Robots, a former robotics moonshot at Google X, sonified a robot's motion such that the robot gave an appearance of dancing while performing any task \cite{cuan2023music}.


\section{Material and Methods}

The objective is to create an 8-hour performance for one human dancer and one industrial robot arm.  We begin by recording video of the human dancer moving in front of a neutral background.  From these videos, we extract vectors of human joint values and smooth these to create sinusoidal functions for each robot joint.  We then use these functions as the initial basis for robot trajectories.  The human dancer returns to the lab and video is recorded of the dancer responding to the robot trajectories.  We extract vectors of human joint values and smooth these to create additional sinusoidal functions for each robot joint.  We then combine these functions to generate novel motion "motifs" that evoke the gestures of human workers such as stirring.  In the process of robochoreography, the artists sequence these motifs with human dance motions and associated music clips.  These motifs are then combined and replayed during the live performance.  We describe these steps in detail in the following sections.

\subsection{OpenPose and Sinusoid Fitting}
\begin{figure}
\begin{center}
\vspace{-0.8cm}
    \includegraphics[scale=0.4]{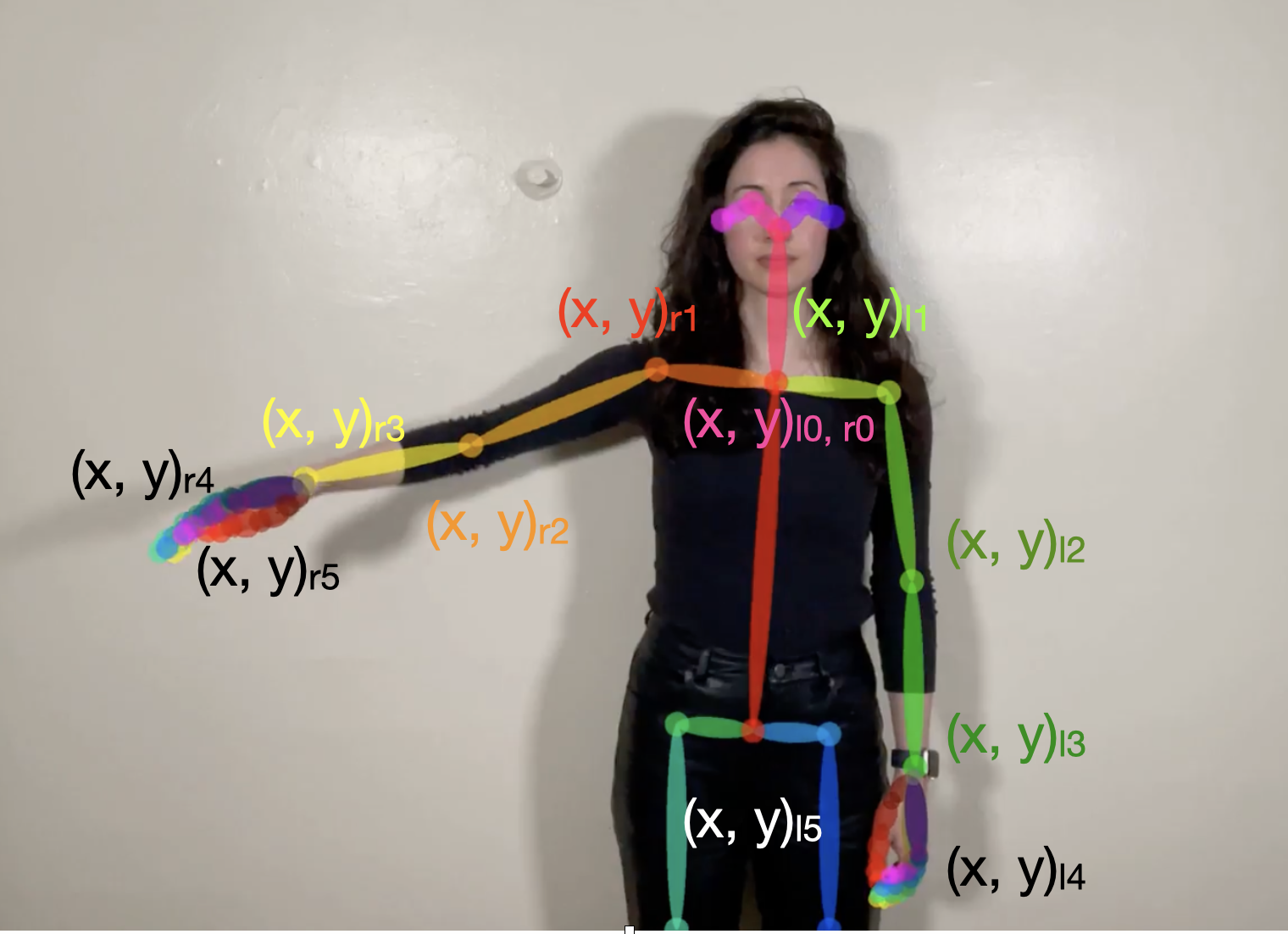}
    \caption{Dancer Catie Cuan with the output of the OpenPose software, indicating automatically segmented kinematic joints. We process these images to extract joint angles over time, then fit these trajectories to sinusoidal functions and use these to program the robot arm.}  
\vspace{-0.8cm}
\end{center}
\label{fig:human_idx}
\end{figure}
We denote a video with $k$ image frames as $v = \{I(1), ..., I(k)\}$. OpenPose, a system for real-time multi-person keypoint detection \cite{cao2017realtime}, processes each frame to generate an overlay image indicating segmented linkages and joints corresponding to the human pose $H_i$.  We developed an algorithm to take as input this sequence of images and processes it to compute $\theta_i$, a corresponding sequence of 6-dimensional vectors of joint angles for the UR5e robot arm.   

\begin{figure}
    \vspace{-0.8cm}
    \centering
    \includegraphics[scale=0.27]{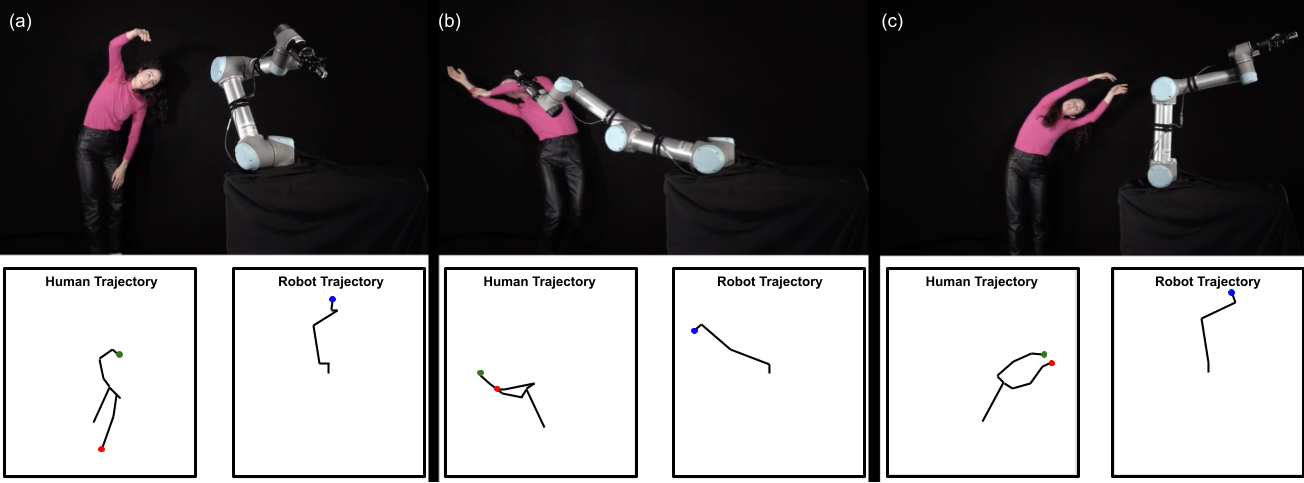}
    \caption{Three snapshots from phase two, where the human dancer returns to the lab to dance with the UR5e robot arm.  The graphs below each image were rendered using matplotlib to illustrate the joint angles extracted from OpenPose.}  
    \vspace{-0.5cm}
\label{fig:openpose}
\end{figure}

For each frame, we extract the following human joints using OpenPose:
\begin{enumerate}
    \item $l_0, r_0:$ Chest, the same coordinate is used in both computations for left and right arms
    \item $l_1, r_1:$ Left and right shoulder
    \item $l_2, r_2:$ Left and right elbow
    \item $l_3, r_3:$ Left and right wrist
    \item $l_4, r_4:$ Left and right hand (averaged using all non-thumb fingers on the hand)
    \item $l_5, r_5:$ Left and right thumb
\end{enumerate}

Given these coordinates, we find the angles for the shoulder, elbow, and wrist joints for the human by computing arc-tangent. The arc-tangent returns an absolute angle, so we further process this by subtracting the angle from the previous joint, similar to how a robot joint is computed.
Using this formula we can compute the shoulder, elbow, and wrist angles.

Hand gestures are often used expressively in performance. We seek to emulate this by computing the angle between the thumb and the wrist as an approximate of the wrist rotation. The previous computation of angle between the hand and the wrist finds the wrist flexion.

We choose the left or right arm that moves the most in our video for continued analysis. To reduce the noise from OpenPose, we smooth the angle trajectories using a 1D blur convolution of size 15. We then apply low-pass frequency filtering in the fourier domain by setting high-frequency bins to 0. Empirically, we find that a threshold of 20 works well to generate trajectories that are interesting and within safety constraints of the robot. We then map these frequencies to the corresponding joints on the UR5e ($H_1$ is mapped to shoulder lift, $H_2$ to elbow, and $H_3$ to wrist 1, and $H_4$ to wrist 3). This leaves the \texttt{shoulder pan} and \texttt{wrist2} joints unmapped. We map low moving frequency sinusoids to these joints to finalize the motion.

\begin{algorithmic}
\For {frame $i$:= 1, ..., n}
\State From this frame, OpenPose extracts pixel cordinates for joints $j \in [0, 1, 2, 3, 4]$ $(x_j, y_j)_{l}, (x_j, y_j)_r$

\State Find the angles for shoulder, elbow, and wrist flexion 
\For {human arm joint $j$:= 0, 1, 2, 3}
\State $H_{l_j} = \text{atan2}(y_{l_{j+1}}-y_{l_j}, x_{l_{j+1}}-x_{l_j})$ 
\State $H_{r_j} = \text{atan2}(y_{r_{j+1}}-y_{r_j}, x_{r_{j+1}}-x_{r_j})$ 
\EndFor
\For {arm joint $j$:= 3, 2, 1}
\State $H_{l_j} \gets H_{l_j} - H_{l_{j-1}}$
\State $H_{r_j} \gets H_{r_j} - H_{r_{j-1}}$
\EndFor

\State Find the wrist rotation using the thumb and wrist positions
\State $H_{l_4} = \text{atan2}(y_{l_{5}}-y_{l_3}, x_{l_5}-x_{l_3})$ 
\State $H_{r_4} = \text{atan2}(y_{r_{5}}-y_{r_3}, x_{r_5}-x_{r_3})$ 

\EndFor

\State Compute the net change of left human arm joints $s_l = \sum_{i=1}^{n-1} = |H_{l_i}-H_{l_{i+1}}|$
\State Compute the net change of right human arm joints $s_l = \sum_{i=1}^{n-1} = |H_{r_i}-H_{l_{i+1}}|$

\If {$s_l \geq s_r$}
\State $H_i \gets H_{l_i}$
\Else
\State $H_i \gets H_{r_i}$
\Comment{We select the arm that moves more for analysis}
\EndIf

\State $H_i \gets \text{Convolve} (H)$ 
\State $f \gets \text{FFT} (H)$  \Comment{Fast Fourier Transform}
\State $f_i \gets 0$ if $|f_i| < \text{threshold}$, else $f_i \gets f_i$ 
\State $H' \gets \text{FFT}^{-1}(f)$ \Comment{Inverse Fourier Transform}

\For {frame $i$:= 1, ..., n}
\State $\theta_i = [\theta_0(t), H_{i0}, H_{i1}, H{i2}, \theta_4(t), H_{i3}]$
\EndFor
\end{algorithmic}

We find $\theta_i$, the joint angles for all 6 of the UR5e's joints. $\theta_0$ and $\theta_4$ are sampled from manually generated sinusoidal equations developed with tools that we detail next section.

\subsection{Sinusoidal Motion Generation}

To control the robot arm, we use sequences of sinusoidal functions of time for each joint of the following form:
\[
\theta_i(t) = A \cos(\omega t + \varphi) + \gamma
\]
For example, stirring is a circular smooth motion whereas hammering requires a strong downward swing and slower lifting reset motion.

\begin{figure}
    \hspace{1cm}
    \includegraphics[scale=0.4]{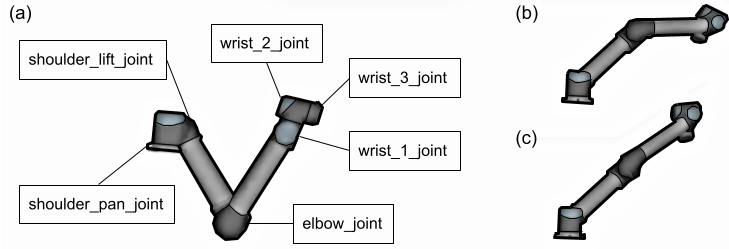}
    \caption{\textbf{Visualization of the UR5e Arm.} 6 joints on the UR5e utilized by the urdfpy simulation, a Python library for loading, manipulating, and exporting URDF files and robot specifications. Figure 4b showcases the UR5e with shoulder$\_$lift$\_$joint at an angle of -$\pi$/4 radians and elbow$\_$joint at $\pi$/4 radians. Figure 4c illustrates the UR5e with shoulder$\_$lift$\_$joint at an angle of -$\pi$/4 radians and elbow$\_$joint at an angle of 0 radians.}  
    \vspace{-0.5cm}
\label{fig:urdfpy}
\end{figure}

When developing any action, we had to first understand the horizontal and vertical range of motion required among its 6 joints as labeled in Figure 4: shoulder$\_$pan$\_$joint, shoulder$\_$lift$\_$joint, elbow$\_$joint, wrist$\_$1$\_$joint, wrist$\_$2$\_$joint, wrist$\_$3$\_$joint. Depending on where the robot should face, we edited the $\gamma$ value of the shoulder$\_$pan$\_$joint which set the angle at which the action is centered around. Knowing that the dancer would perform stage right of the UR5e, the shoulder$\_$pan$\_$joint is centered around 0 radians to face the back of stage, $\pi$/2 to directly face her, $\pi$ to face the audience, or 3$\pi$/2 to face stage left by setting $\gamma$ to these values. The amplitude of the movement was determined by the value of A, which ranged from 0 to $\pi$/2 radians depending on the joint and motion requirements. Joints such as shoulder$\_$pan$\_$joint and shoulder$\_$lift$\_$joint often had larger amplitudes since they were capable of extending the UR5e's reach further than the others. When experimenting with motions, we found that utilizing $e^{-Bx}$ instead of a constant value of A, creates a gradually increasing or diminishing effect on motion amplitude. We used this effect in a bartender sequence to mimic the action of pouring a drink, slowly decreasing the height to suggest pouring the drink.

Each joint had its own coordinate frame relative to the shoulder$\_$pan$\_$joint. To achieve a desired effect, we had to understand each joint's required position relative to the previous joint. For example, the angles of shoulder$\_$lift$\_$joint can operate on the z-axis of the coordinate frame parallel to the floor since its previous joint is shoulder$\_$pan$\_$joint which operates on an xy-plane parallel to the floor. However, the sinusoidal equation for the elbow$\_$joint must be carefully adjusted based on the position of the shoulder$\_$lift$\_$joint. While setting the shoulder$\_$lift$\_$joint to an angle of 0 would place the first part of the UR5e arm parallel to the floor, setting the elbow$\_$joint angle to 0 while shoulder$\_$lift$\_$joint is at -$\pi$/4, would result in the position illustrated by Figure 4c rather than 4b.  

Experimentation was key to determining these values. Before testing any actions on the UR5e, we visualized them using the urdfpy graphics library \cite{urdfpy}. The simulation takes six arrays of joint angles in radians as inputs to correspond to each of the six joints on the UR5e. These arrays outlined the trajectory that the urdfpy \texttt{animate} function executes, simultaneously setting each joint according to the values in its array. This allows us to visualize the combination of joint trajectories and fine-tune angles and speeds accordingly by evaluating.

\begin{figure}
    \hspace{0.4cm}
    \includegraphics[scale=0.21]{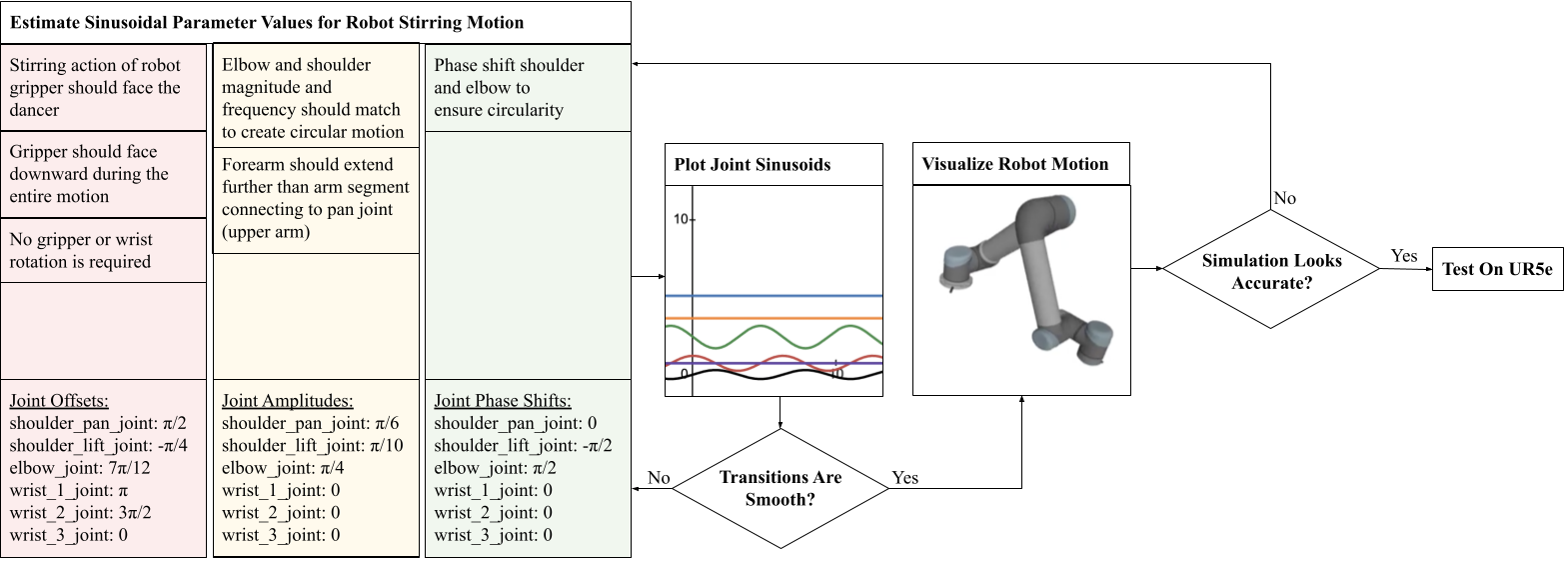}
    \caption{\textbf{Methodology Behind Stirring Action Development} Example of pipeline for setting sinusoidal parameter values for robot motions that suggest the robot arm is stirring a pot of soup.}  
    \vspace{-0.5cm}
\label{fig:stirring_pipeline}
\end{figure}

The process of developing a motif is illustrated in Figure 5 for a motion of the robot stirring a large pot of soup.

We visualize all the motions in simulation to confirm that each action would occur at a desirable speed. It was crucial for the safety of the human dancer and robot hardware that all potential safety hazards were fixed in simulation before transferring them to the UR5. Hazards typically referred to sudden jolts at high speeds which could easily damage internal motors, as well as robot self-collisions. Sudden jolts would occur if there was a large gap between two joints' angular positions. Self-collisions were a result of a miscalculated or overcompensated joint pathway leading one or more joints to crash into another part of the arm while executing their trajectory. In extreme cases, the UR5e was able to activate its emergency stop, allowing us to reset its position and adjust the above constants to address the issue following the aforementioned equation adjustment process.

\subsection{Move to Teach, Replay (Force Sensing)}

We also include live interactive sections between the robot and the dancer. The UR5e's \texttt{teachMode} enables a "Zero-G" mode, where the robot is compliant to all external forces, and can be moved by the dancer for a period of time. This allows the robot to record its motions while being moved by hand, and then replay the motion. To maximize fluidity, we recorded and replayed joint angles at 500Hz, the maximum UR5e control frequency.

\begin{figure}[h]
\begin{center}
    \includegraphics[scale=0.4]{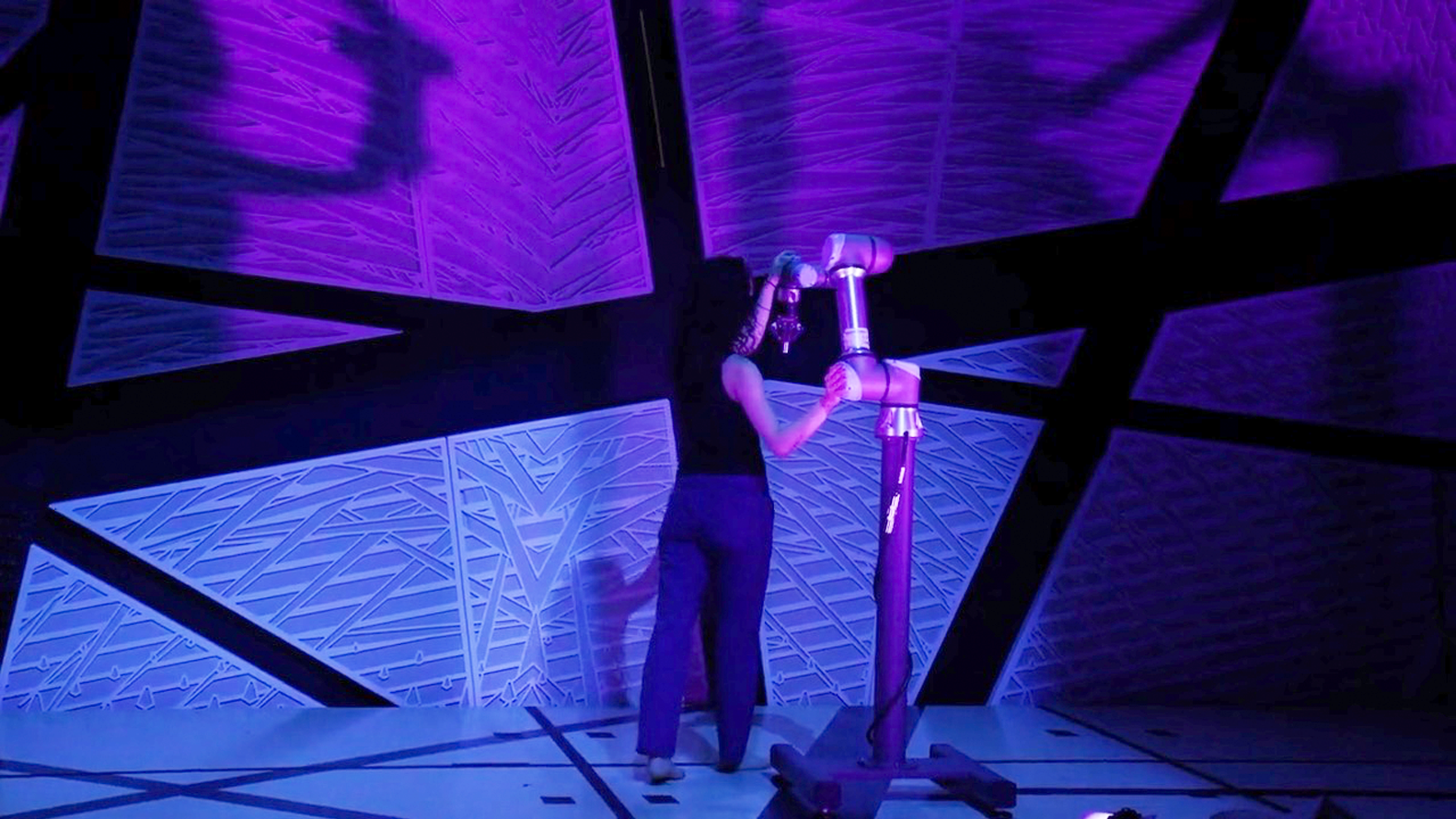}
    \caption{Live performance photo, National Sawdust Theater, 2023.}  
\vspace{-0.8cm}
\end{center}
\label{fig:interaction}
\end{figure}

The UR5e's safety mechanism forbids the robot to transition from teach mode to regular position/ joint control mode when there are significant amounts of external forces on the robot. This posed a challenge to the performance: Catie needed to stop interacting with the robot before the transition happens to minimize robot protective stops, but in a live performance, it was difficult to indicate upcoming transitions in a non-intrusive way, especially since the accompanying music was not designed such that each motif has its own individual score.

To address this challenge, we used the UR5e's \texttt{forceMode} feature. Originally designed to aid in tasks such as drill pressing or screwing, this function enables the robot to be "partially compliant", and continuously apply a force in some direction with some damping. The damping parameter specifies how quickly the robot returns to stationary when a force is removed. A value of 1 is full damping, so the robot will decelerate quickly if no force is present. A value of 0 is no damping, here the robot will maintain the speed. We take advantage of this damping parameter and add an section between the compliant and rigid sections of the robot at a damped value of 0.2. After a fully compliant \texttt{teachMode} section, the robot enters a section of \texttt{forceMode}, during which the it is compliant but heavily resisted forces applied by Catie, enabling seamless transitions

For a live performance, we also designed sections that can played after waiting for a cue. For example, the concluding bow. A firm push on the robot suffices as an effective and visually pleasing prompt. The force features of the UR5e once again proved helpful. After entering a waiting stage, the robot recorded the readings given by the force sensor at the end effector. When a large force is detected, the robot executes the proper motion. The UR5's force sensor can be noisy, and we apply a moving average to the noise reading to detect this tap: at initialization, the running average is set to 0, and we keep track of the past 10 readings. When this average exceeds our threshold (20 N). Below we display the raw values as well as the computed value in rehearsal.

\begin{figure}[H]
\begin{center}
\vspace{-0.5cm}
    \includegraphics[scale=0.4]{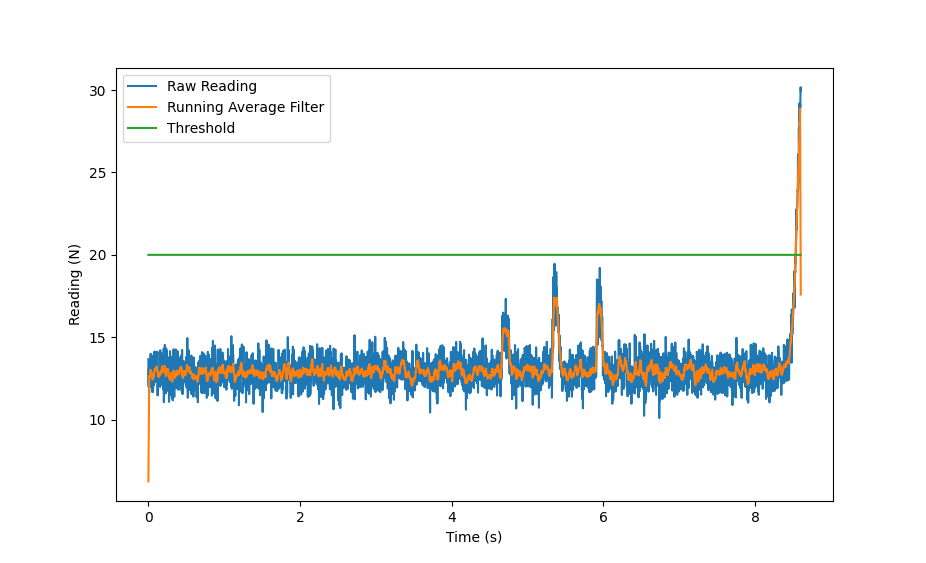}
    \caption{The raw signal vs the running average detected on the end effector. The robot did not react to 3 accidental bumps against it and correctly identify the cue when intentionally pushed at 9 seconds.}  
\vspace{-0.8cm}
\end{center}
\label{fig:force_sensing}
\end{figure}

\subsection{Sound Design and Robot Choreography}

Sound design was integral to the performance, as it maintained artistic momentum and kept the audience's attention. The eight hour performance was conducted to both original and preexisting music. The music varied in genres, tempo, and instrumentation. The choreographer considered it important to vary the music in a cyclical way, so that high tempo, high energy music was followed by slower tempo, lower energy music. Overall, no songs were repeated throughout the eight hour performance. Seven pieces of original music were written by composer Peter Van Straten.

A sampling of included genres and composers are mentioned below:
\begin{itemize}
    \item Electronic: Four Tet, Son Lux, Szymon
    \item Rock: The Clash, David Bowie, The Rolling Stones
    \item Classical: Wolfgang Amadeus Mozart, Frédéric Chopin
    \item Contemporary Classical: Caroline Shaw, Meredith Monk, Max Richter
    \item Jazz: Count Basie Orchestra, Stan Getz, George Gershwin
    \item Rhythm \& Blues: Solange Knowles, Emily King
\end{itemize}

As noted above, we create a series of ``Movement motifs'' that repeat at various points in the performance to support the concept of illustrating an American workday. For example, one movement motif was a bartending scene. In this case, the robot performed the shaking action as if mixing a drink with a shaker.  

Pairing the movement motifs with the music took several rounds of iteration. This was because the robot transitions into and out of prerecorded motions and moves to teach mode. At the end of a motif, we design a heuristic that determines whether the next motion may lead to possible collisions. Specifically, we filter the joint angles to detect if the robot is at risk of self-collision between the wrist and the elbow. If this is the case, we enter into a safe transition sequence to interpolate smoothly and safely into the next motif section. In addition, the musical and movement transitions needed to make artistic sense. For example, if several literal physical labor motifs all occur in a row, they are followed by more abstract robot motions and free form dancing. This was to remove the expectation that a linear narrative was unfolding on stage.
\begin{figure}[H]
\begin{center}
    \includegraphics[scale=0.4]{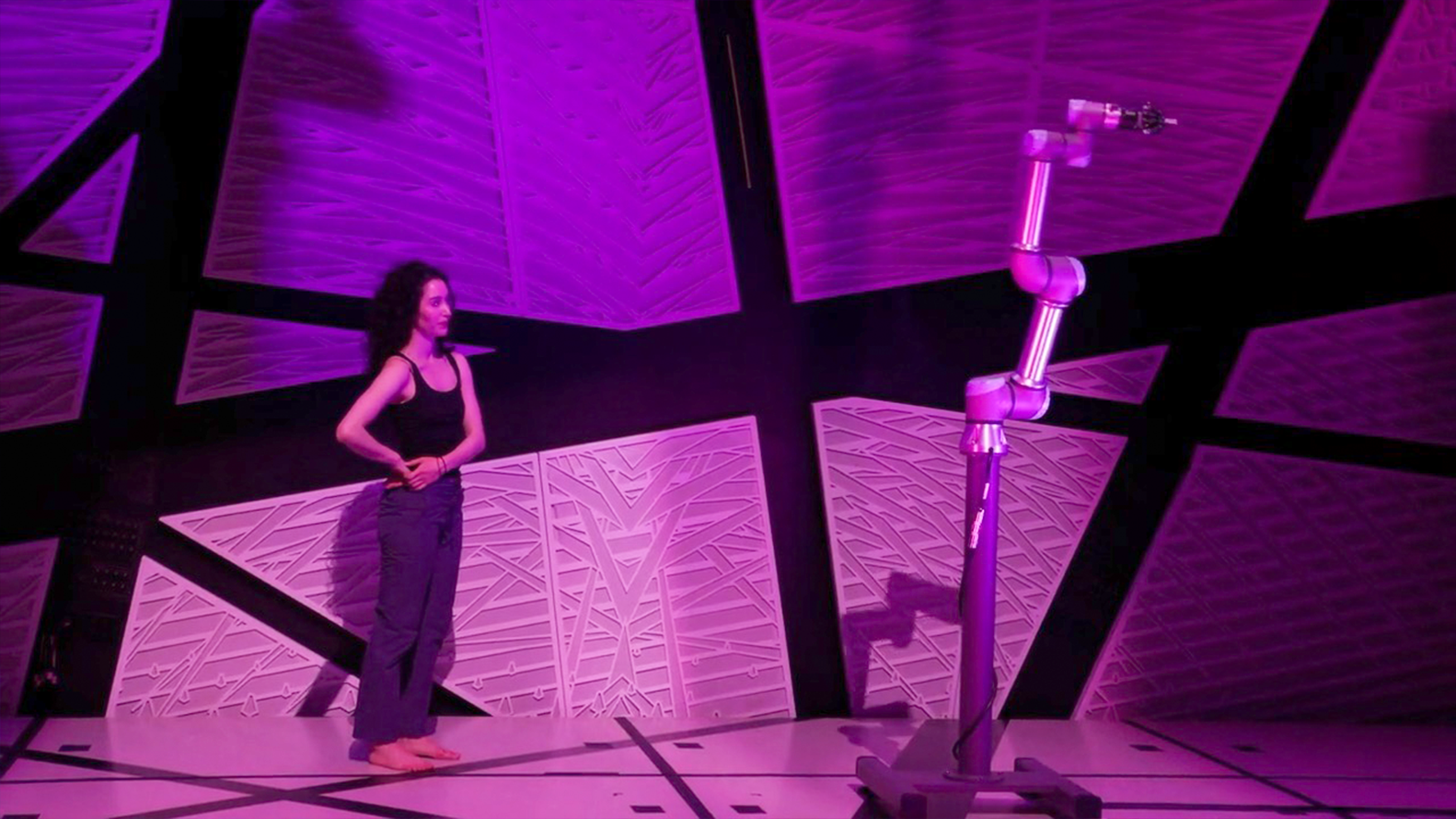}
    \caption{Live performance photo, National Sawdust Theater, 2023.}  
\end{center}
\vspace{-0.8cm}
\label{fig:running}
\end{figure}

\subsection{Live Cuing}

As the human dancer's motion motifs and the corresponding robot motion motifs and the paired musical selections (with timing information) were changing until just before the performance, we used Google Sheets to keep track of the updating sequence of motifs and music.  However, as this format was too small to be visible to the dancer when performing on stage, we used custom Applescript code to convert the .xls file into a set of Apple Keynote slides -- one for each motif -- which were then displayed via zoom on a small laptop downstage to cue the human dancer (Cuan) in real time.  These slides were also used by the lighting director (Goldberg) to plan and request stage lighting effects: color, intensity, and location of digital LED lights, in real time during the 8 hour performance.

\vspace{-0.3cm}
\section{Results}
\vspace{-0.3cm}
The performance on December 16, 2023, was attended by approximately 600 individuals over eight hours. The  robot successfully completed the performance without needing to be replaced or significantly debugged. In a few cases, it timed out and needed to be reset, but it largely ran a collection of pre-recorded and live-cued actions for eight hours. The human dancer likewise performed the entire eight hour performance without significant injury, only taking two 15 minute breaks while the existing audience exited and was replaced by the next audience.

One audience member noted, ``One of the most impressive performance pieces I have seen in a very long time and a real game-changer. One of those rare cultural moments where you have goosebumps whilst being transfixed by a performance between a dancer and machine, of such deep originality, profound poetry, insight, and integrity, that your heart and mind can't stop thinking about  and feeling it for  many days afterwards and beyond.''

In Forbes Magazine \cite{wolff_teaching}, ``Some of the most astonishing and dramatic moments in “Breathless” come when Cuan slides up to the robot arm, her head just inches from one of its large metal joints. She reaches in, almost caressing the arm, and gently pushes and pulls it through a sequence of movements. Then she lets go, and the robot replays those exact motions as she dances alongside—a 21st century pas de deux.'' Audience members noted the hypnotic nature of the performance, running for eight hours with movement and sound variation.

\vspace{-0.5cm}
\begin{figure}[H]
\begin{center}
    \includegraphics[scale=0.25]{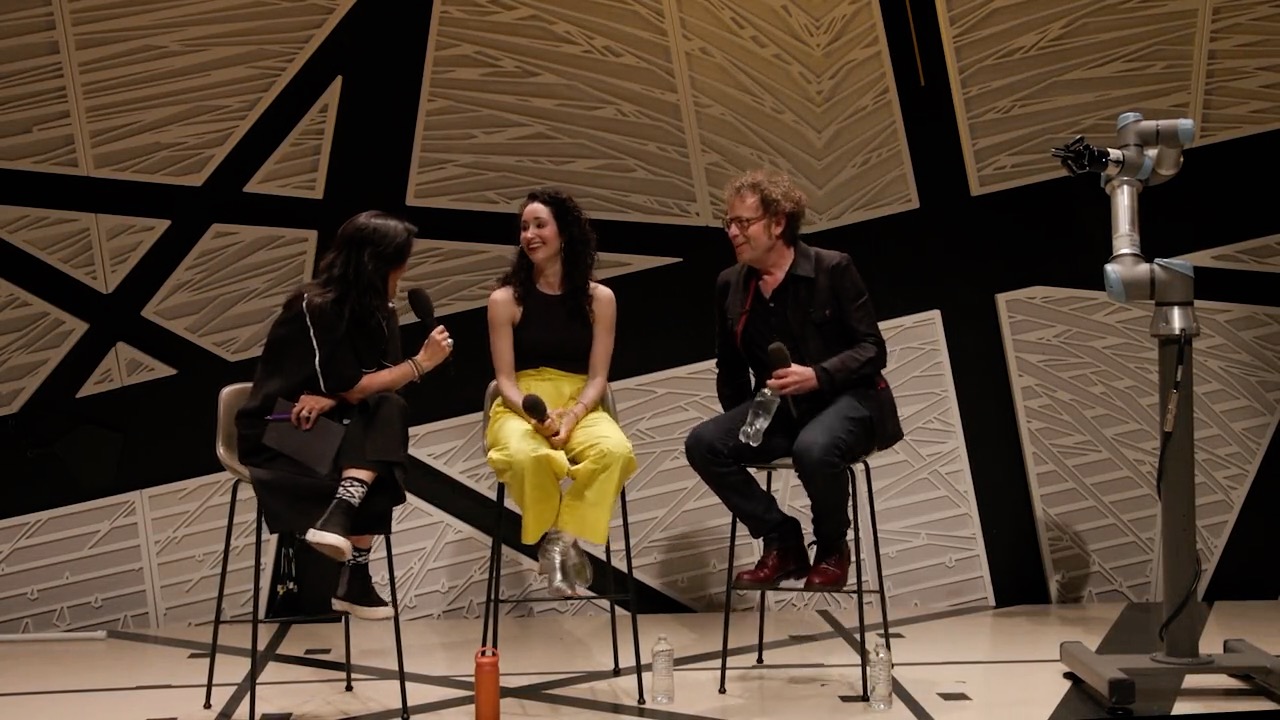}
    \caption{The interview after the performance featuring Elena Park, Catie Cuan, and Ken Goldberg}  
\end{center}
\vspace{-0.8cm}
\label{fig:interview}
\end{figure}

\newpage

\section{Limitations and Future Work}

The performance ran smoothly and was very well-received by the audience, who appreciated the stamina and nuanced physical performance of the human dancer in contrast to the smooth but inherently mechanical motions of the robot arm.  The UR5e robot and Universal Robots leadership and technical team were a pleasure to work with.  We experienced a major scare when the robots were substantially delayed during shipping, requiring us to rent a truck, retrieve them from the back of a Fed-Ex delivery truck, and drive them from New Jersey to Manhattan less than 24 hours before the performance began.  The existing software system is complex to operate, requiring 3 technical personnel in addition to a stage manager and lighting technician, as well as the human dancer to actively participate throughout the 8-hour performance.  In future work we plan to streamline the software and explore additional motion motifs.  The West Coast premier of the performance is scheduled for 12 Dec 2024.

\section{Acknowledgements}

The authors graciously acknowledge National Sawdust artistic curator Elena Park, line producer David Imani, composer Peter van Straten, Co-Commissioners Jacob's Pillow NS NationalSawdust+, Co-Sponsor + Development Lab AUTOLab at UC Berkeley, Rehearsal Space loan by Lisa Wymore,  Custom robot stand by Jake Hinkemeyer at ai.motion.com, Idea and Program for Fourier Processing by Will Panitch, Fiscal Sponsor + Development Lab Jacob's Pillow + the Dancerly Intelligences Pillow Lab, and the loan of two UR5e Robot Arms by Anders Billesø Beck, Phillip Grambo, of 
Universal Robots and the crew at National Sawdust: Jeff Tang, LeeAnn Rossi, Alyana Vera, Alex Barnes, Bruce Steinberg, and Marielle Iljazoski.

\printbibliography

\end{document}